\documentclass{article}

\PassOptionsToPackage{numbers, compress}{natbib}

\usepackage[preprint]{neurips_2026}

\usepackage[utf8]{inputenc} 
\usepackage[T1]{fontenc}    
\usepackage{hyperref}       
\usepackage{url}            
\usepackage{booktabs}       
\usepackage{amsfonts}       
\usepackage{nicefrac}       
\usepackage{microtype}      
\usepackage[most]{tcolorbox}
\usepackage{xcolor}         
\usepackage{graphicx}
\usepackage{subcaption}
\usepackage{algorithm}
\usepackage{algpseudocode}

\usepackage{multirow}
\usepackage{colortbl}
\usepackage[table]{xcolor}
\newcommand{\resulttabfont}{\small}
\newcommand{\resulttabstretch}{\renewcommand{\arraystretch}{1.08}}

\definecolor{lightblue}{RGB}{224,235,255}

\title{OGPO: Offline Goal-conditioned Policy Optimization for  Recoverable Vision-Language-Action Models}

\author{
Xule Gao$^{1,2,3}$ \quad
Xi Wang$^{1,2,3}$ \quad
Zehua Zang$^{4}$ \quad
Rui Wang$^{1,2}$
\\[0.3em]
Changwen Zheng$^{1,2}$ \quad
Chuxiong Sun$^{1,2,\dagger}$
\\[0.5em]
$^{1}$ National Key Laboratory of Space Integrated Information System \\
$^{2}$ Institute of Software, Chinese Academy of Sciences \\
$^{3}$ University of Chinese Academy of Sciences \\
$^{4}$ Wangxuan Institute of Computer Technology, Peking University \\
\texttt{gaoxule24@mails.ucas.ac.cn} \\
$^{\dagger}$ Corresponding author: \texttt{chuxiong2016@iscas.ac.cn}
}

\begin{document}

\maketitle

\begin{abstract}
Vision-language-action (VLA) models have shown strong promise in robotic manipulation, yet their post-training still predominantly relies on trajectory-level supervised imitation over robot demonstrations. While effective for learning feasible actions, this paradigm supervises policies to reproduce expert actions at each step, biasing them toward demonstration-specific action paths and providing limited guidance for recovering toward task-progress states once execution deviates from demonstrated trajectories. To address this limitation, we propose \textsc{OGPO}, an \underline{O}ffline \underline{G}oal-conditioned \underline{P}olicy \underline{O}ptimization framework for recoverable VLA post-training. Instead of treating offline demonstrations merely as step-wise action labels, OGPO adaptively relabels trajectories with semantic key states and constructs goal-conditioned process rewards to optimize the policy for reaching these task-progress states. In this way, OGPO shifts VLA post-training from trajectory-level action imitation to key-state reachability learning, enabling policies to acquire a more robust ability to complete goal states from feasible off-demonstration states without requiring additional online interaction. Experiments on LIBERO and MetaWorld show that OGPO consistently improves VLA policy performance over standard supervised fine-tuning. Moreover, evaluations on LIBERO-Plus and our proposed LIBERO-ReAct, a mid-execution perturbation benchmark, further demonstrate that OGPO achieves stronger robustness and goal-directed recoverability under distribution shifts and recoverable off-demonstration states.
\end{abstract}

\section{Introduction}
\begin{figure}[t]
    \centering

    \begin{subfigure}[t]{0.48\linewidth}
        \centering
        \includegraphics[width=\linewidth]{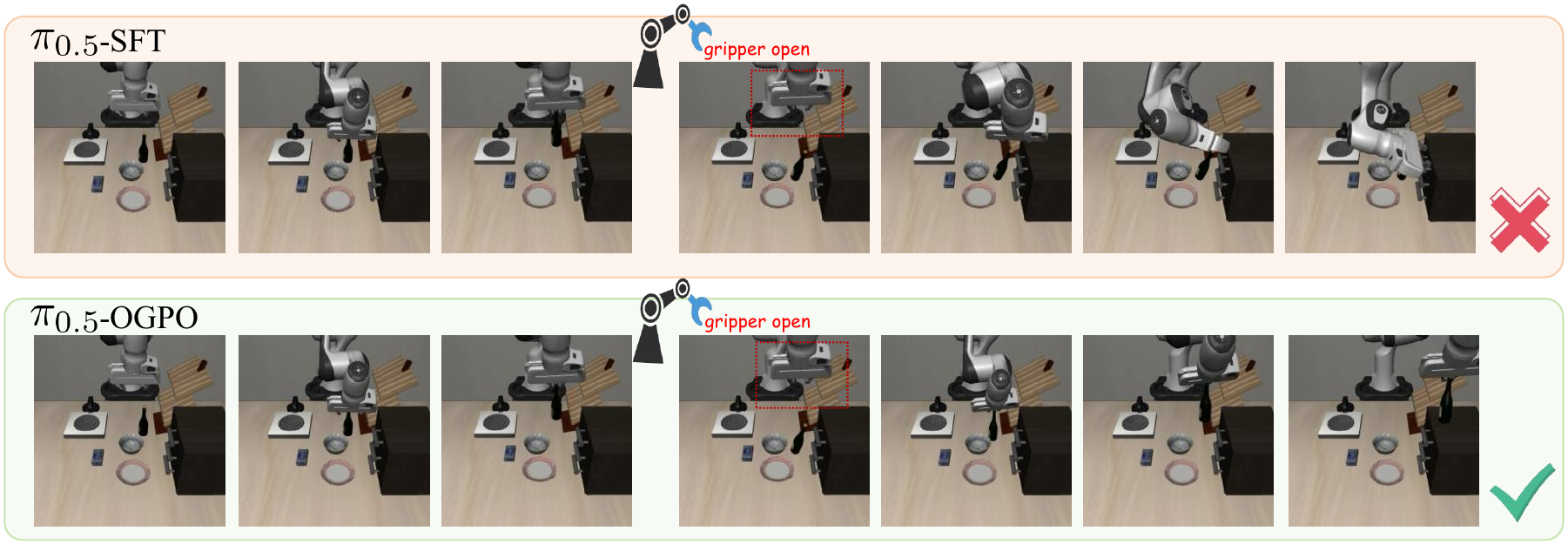}
        \caption{Put the wine bottle on top of the cabinet.}
        \label{fig:libero_spatial_e}
    \end{subfigure}
    \hspace{0.01\linewidth}
    \begin{subfigure}[t]{0.48\linewidth}
        \centering
        \includegraphics[width=\linewidth]{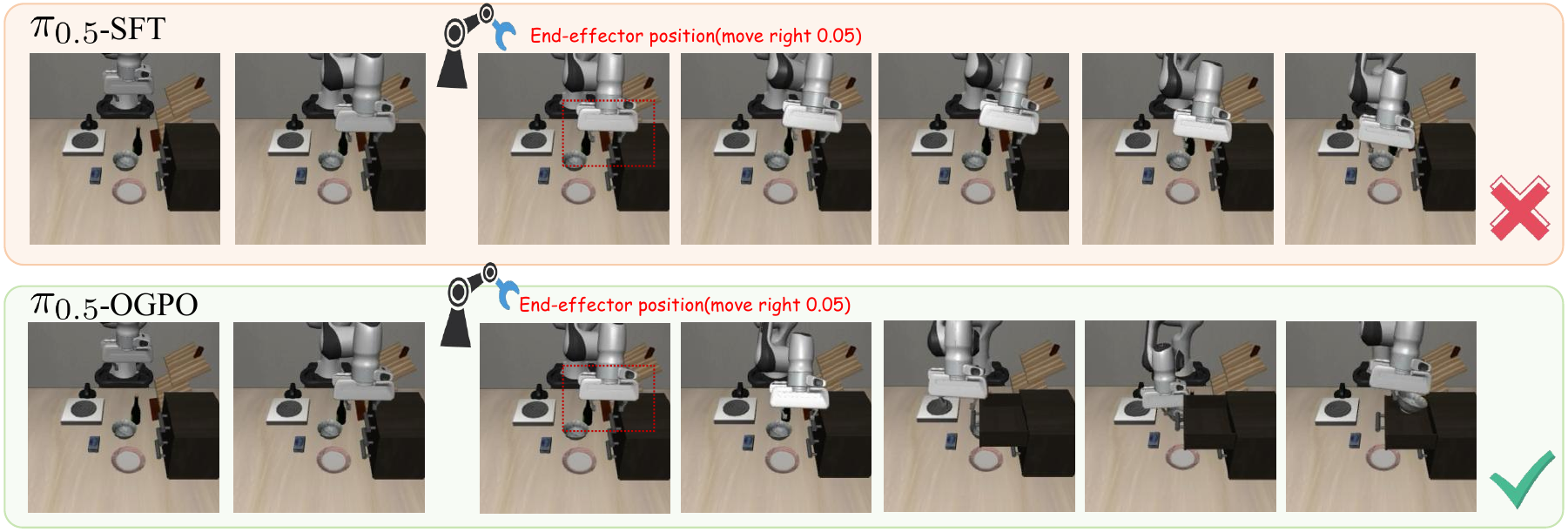}
        \caption{Open the top of drawer and put the bowl inside. }
        \label{fig:libero_goal_e}
    \end{subfigure}
    \caption{We design two representative perturbation cases to evaluate the robustness and goal-directed recoverability of offline-trained VLA policies under deviations from demonstration trajectories. In case (a), the gripper is forcibly opened after grasping the object, while  position perturbation is applied when the end effector approaches the drawer. The results show that $\pi_{0.5}$-SFT fails to complete the task, whereas $\pi_{0.5}$-OGPO successfully recovers toward the goal state and finishes the task, demonstrating stronger recoverability under execution-time deviations.}
    \label{fig:case-pertuation}
\end{figure}

Vision-language-action (VLA) policies have shown strong potential for language-conditioned robotic manipulation by scaling supervised action prediction on large-scale robot datasets~\cite{rt1,rt2,openvla}. Most existing VLA policies are adapted to downstream tasks through supervised fine-tuning (SFT), where the model is optimized to predict expert actions from visual observations and language instructions on demonstration trajectories~\cite{openvlaoft,octo,pi0,pi05}. Although this paradigm is simple and stable, it treats demonstrations primarily as step-wise action labels for trajectory-level imitation. As a result, policies trained purely with SFT tend to reproduce demonstration-specific action paths and remain inherently bounded by the coverage and quality of demonstration data. As shown in Figure~\ref{fig:case-pertuation}, when execution deviates from demonstrated trajectories, such policies may suffer from compounding errors in long-horizon tasks and lack explicit guidance for recovering toward task-progress states~\cite{dagger}.

Recent studies have explored reinforcement learning (RL) to further improve pretrained VLA policies beyond supervised imitation, including iterative online RL fine-tuning~\cite{ire-vla}, PPO/GRPO-style policy-gradient optimization~\cite{vla-rl,simplevlarl,pirl}, residual policy learning~\cite{selfimproving}, and RL-based policy distillation~\cite{rpd}. By optimizing task-level rewards, RL can in principle encourage goal-directed behaviors that are not directly captured by step-wise action imitation. However, existing RL-based VLA post-training methods typically require costly environment rollouts and suffer from limited sample efficiency, making them difficult to scale for real-world robotic post-training.

In this work, we study whether VLA policies can acquire goal-directed recoverability purely from offline robot data, without requiring additional online interaction. We define this ability as recovering toward task-progress key states when execution deviates from demonstrated trajectories. Our key insight is that offline demonstrations should not be used merely as action sequences to imitate; they also contain implicit evidence about which task-progress states are reachable and which behaviors move the policy toward them. Based on this insight, we propose \textsc{OGPO}, an offline goal-conditioned policy optimization framework for recoverable VLA post-training. Specifically, we retain SFT as a lightweight action prior to preserve low-level action feasibility and language-action alignment, while introducing an offline goal-conditioned RL objective to learn key-state reachability from existing trajectories. 
To turn offline data into goal-reaching supervision, \textsc{OGPO} adaptively relabels trajectories with task-progress key states and constructs goal-conditioned process rewards to guide both policy optimization and value estimation. Instead of supervising the policy to imitate every action along the demonstration path, this objective trains the policy to reach task-progress states under goal-conditioned rewards. In this way, \textsc{OGPO} shifts VLA post-training from trajectory-level action imitation to key-state reachability learning, enabling policies to improve goal-directed recoverability under feasible off-demonstration deviations while avoiding the high interaction cost of online RL. 

We evaluate OGPO on LIBERO~\cite{libero} and MetaWorld~\cite{metaworld}, and further test robustness and recoverability on LIBERO-Plus~\cite{liberoplus} and our proposed LIBERO-ReAct benchmark with mid-execution perturbations. The results show that OGPO consistently improves over SFT baselines, especially on long-horizon and difficult tasks, while achieving stronger robustness and goal-directed recoverability under distribution shifts and execution-time deviations. Our contributions are threefold. 
\begin{itemize}
    \item First, we formulate VLA post-training from the perspective of goal-directed recoverability and identify the limitation of trajectory-level SFT in recovering toward task-progress states under execution deviations.
    \item Second, we propose OGPO, an offline goal-conditioned policy optimization framework that adaptively relabels fixed robot trajectories with task-progress key states and constructs goal-conditioned process rewards, allowing the policy to learn and optimize goal-directed behaviors beyond step-wise action imitation.
    \item Third, we conduct extensive experiments on standard manipulation, distribution-shift, and mid-execution perturbation benchmarks, demonstrating that OGPO consistently improves task performance, robustness, and goal-directed recoverability without requiring additional online interaction.
\end{itemize}

\section{Related Works}

VLA post-training aims to adapt pretrained vision-language-action policies to downstream robotic manipulation tasks while preserving their general action priors and improving task-specific performance. 
Existing approaches mainly follow two directions: SFT and RL. 

\subsection{Supervised Fine-tuning for VLA Policies}
SFT has become a dominant post-training strategy for adapting pretrained VLA policies to downstream robotic tasks. In this paradigm, the model is fine-tuned on expert demonstration trajectories by minimizing a supervised action prediction loss conditioned on visual observations and language instructions. Existing VLA models instantiate this paradigm with different action representations and prediction objectives. OpenVLA~\cite{openvla} discretizes continuous robot actions into action tokens and optimizes the policy with a next-token prediction objective using cross-entropy loss. Octo~\cite{octo} predicts continuous action chunks through a diffusion-based action head trained with a diffusion loss. $\pi_0$~\cite{pi0} and $\pi_{0.5}$~\cite{pi05} further adopt flow-matching objectives for continuous action-chunk prediction. Recent studies such as OpenVLA-OFT~\cite{openvlaoft} improve SFT-based adaptation by introducing continuous action representations, action chunking, parallel decoding, and regression-based imitation objectives. Despite these advances, SFT remains inherently constrained by the coverage and quality of demonstration data: it primarily learns to reproduce expert actions along demonstrated trajectories, and thus provides limited support for generalizing to unseen states, handling execution perturbations, and mitigating long-horizon compounding errors~\cite{dagger}.

\subsection{Reinforcement Learning for VLA Post-training}
Recent works have explored RL as an additional post-training stage after SFT. By optimizing task-level rewards through online rollouts, RL allows VLA policies to move beyond the behavior distribution of expert demonstrations and potentially acquire more robust behaviors under distribution shift~\cite{rlwhat}. For example, iRe-VLA~\cite{ire-vla} alternates between online RL and supervised learning to improve policy performance while maintaining training stability. Other methods apply PPO- or GRPO-style policy-gradient optimization to VLA models~\cite{vla-rl,simplevlarl,pirl}, aiming to enhance long-horizon decision-making and task success through outcome-driven rewards. In addition, residual RL methods improve pretrained VLA policies by learning lightweight corrective policies around frozen generalist models~\cite{selfimproving}, while RL-based policy distillation methods leverage VLA teachers or RL experts to obtain more specialized policies~\cite{rpd}. More recent studies further explore simulator- or world-model-based reinforcement fine-tuning to reduce the cost of direct real-world interaction~\cite{worldenv,vlarft,wmpo,wovr}. Although these methods demonstrate the promise of RL for improving VLA policies beyond SFT, they typically rely on online rollouts, real-world interaction, or simulator-based policy evaluation, making them costly and sample-inefficient for long-horizon robotic tasks.

\paragraph{Our distinction.}
In this work, we combine the benefits of SFT and RL in a fully offline VLA post-training framework. 
Different from prior methods that rely mainly on action imitation, online rollouts, or simulator-based policy evaluation, OGPO extracts goal-reaching and failure-aware supervision directly from offline robot data. 
This enables VLA policies to improve goal-directed recoverability without additional environment interaction. 
To the best of our knowledge, OGPO is the first framework that integrates SFT regularization with offline goal-conditioned reinforcement learning for recoverable VLA post-training.

\section{Method}
In this section, we present OGPO, an offline goal-conditioned policy optimization framework for recoverable VLA models. OGPO operates in three stages. First, it relabels offline trajectories into key-state-centered goal-reaching segments, thereby exposing the task-progress structure hidden in demonstrations. Second, it converts segment-level completion labels into goal-conditioned process rewards, which provide dense learning signals for value estimation. Third, it performs offline advantage-weighted policy optimization with SFT regularization, enabling the policy to improve toward goal-reaching behaviors while preserving feasible action priors.

\begin{figure}[t]
    \centering
    \includegraphics[width=\linewidth]{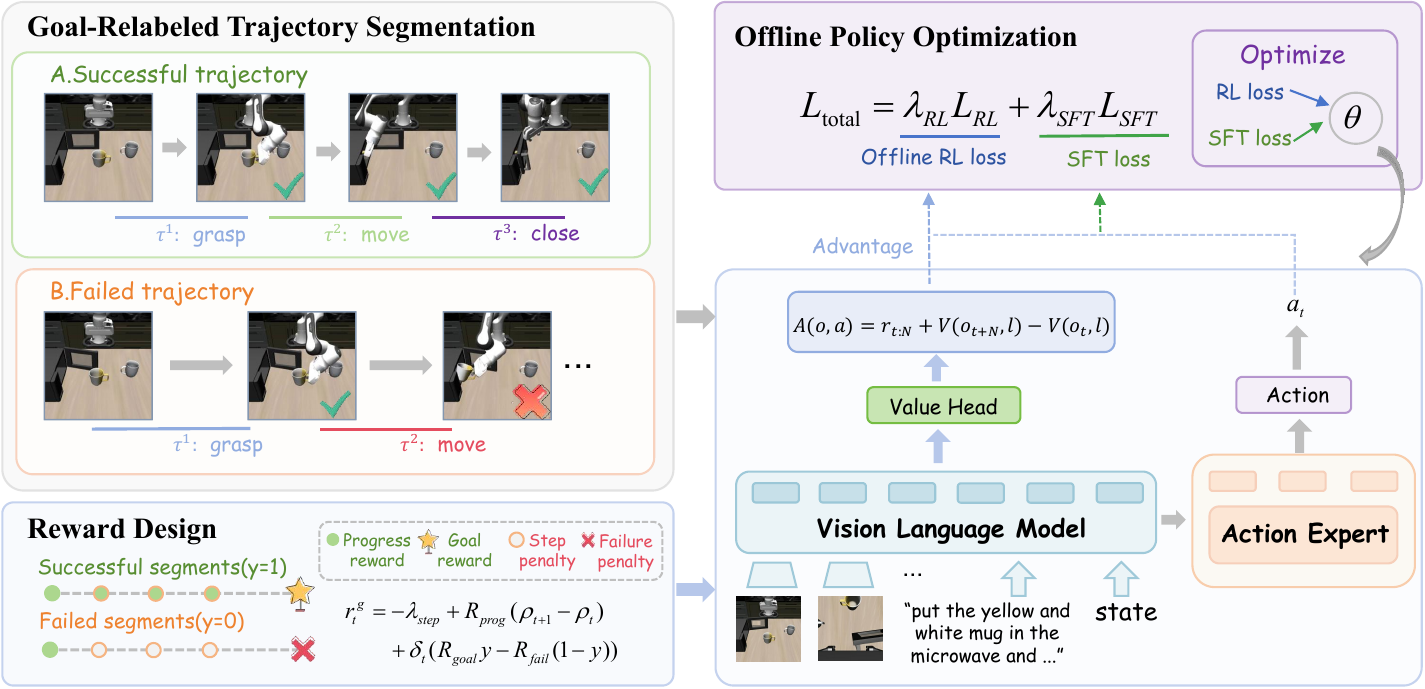}
    \caption{Offline Goal-conditioned Policy Optimization(\textsc{OGPO}) framework.}
    \label{fig:framework}
\end{figure}

\subsection{Goal-Relabeled Trajectory Segmentation}
To discover the task-progress structure embedded in offline trajectories, OGPO first segments each trajectory according to  gripper state change. Given an expert demonstration trajectory \(\tau\):
\begin{equation}
\tau = \left\{\{(o_t,a_t)\}_{t=1}^{T}, l\right\}
\label{eq:demo_trajectory}
\end{equation}
where \(o_t\) and \(a_t\) denote the observation and action at timestep \(t\), and \(l\) is a natural language instruction describing the task carried out in the demonstration. We assume that each trajectory \(\tau\) is composed of a sequence of task-progress stages. The objective of trajectory segmentation is to split the trajectory into non-overlapping temporal segments \(\{\tau^1,\tau^2,\ldots,\tau^M\}\), each corresponding to a task-progress stage. Following Chen et al.~\cite{atomic} and  Zhu et al.~\cite{deco}, we realize such segmentation based on keyframes containing the change of gripper state. Despite its simplicity, gripper state change is a robust indicator of critical physical interaction, which can be viewed as a signal of the end of one stage and the beginning of the next.



After segmentation, we perform goal relabeling for each segment. Specifically, we first assign each segment $\tau^m$ a semantic local goal $g^m$ based on the task instruction, gripper transitions, and observed state changes, as detailed in Appendix~\ref{app:goal_assignment}. We then assign each segment a binary completion label
\(y^m=\mathbb{I}[\tau^m \text{ achieves } g^m]\), using different labeling procedures for successful and failed trajectories:

\begin{itemize}
    \item For a successful trajectory, each valid segment is treated as completing its assigned local goal and is therefore labeled with $y^m=1$. The segment-end observation is regarded as the corresponding achieved task-progress key state.
    \item For a failed trajectory, the task-level outcome does not directly indicate which local goals have been completed, as meaningful intermediate progress may occur before the eventual failure. We therefore introduce an offline VLM-based annotator to evaluate each segment and assign a binary completion label $y^m$. Details of the completion labeling are provided in Appendix~\ref{app:completion_labeling} and ~\ref{app:vlm_prompt}.
\end{itemize}

After relabeling, we obtain a goal-relabeled offline dataset:
\begin{equation}
\widetilde{\mathcal{D}}=\{(o_t,l,a_t,g_t,y_t)\}.
\label{eq:relabeled_dataset}
\end{equation}
Through this process, we obtain a diverse goal-relabeled dataset that contains both achieved and unfulfilled task-progress key states, providing a solid foundation for subsequent VLA policy optimization.

\subsection{Designing Goal-Conditioned Process Rewards for Value Learning}

Directly constructing a goal-conditioned policy on top of a pretrained VLA model would require architectural modifications and may interfere with the pretrained vision-language representations. To avoid this, we use goal-conditioned process rewards as training-time signals for offline policy optimization. This design preserves the original VLA policy structure and instruction-conditioned action interface, while still enabling the policy to acquire goal-directed behaviors by optimizing toward task-progress key states. Concretely, we construct goal-conditioned process rewards from the relabeled key states and completion labels. The relabeled goal \(g_t\) is used only to annotate rewards and define segment-level goal progress during training; it is not introduced as an additional input to either the policy or the value function. 

For each relabeled segment \(\tau^m=\{(o_t,l,a_t,g_t,y_t)\}_{t=s_m}^{e_m}\), where \(s_m\) and \(e_m\) denote the start and end time steps of the segment, we define a normalized progress variable:
\begin{equation}
\rho_t^m
=
y_t \cdot
\frac{t-s_m}{e_m+1-s_m},
\qquad
t\in [s_m,e_m+1].
\label{eq:progress_variable}
\end{equation}
When \(y_t=1\), \(\rho_t^m\) increases from \(0\) to \(1\) along the segment, providing dense progress reward for a completed goal. When \(y_t=0\), \(\rho_t^m=0\) for the entire segment, so failed segments do not receive progressive positive rewards.

For each transition \(t\in[s_m,e_m]\), we use a segment-end indicator \(\delta_t^m=\mathbb{I}[t=e_m]\), which equals 1 only at the final transition of the segment and 0 otherwise. The reward is then annotated as:
\begin{equation}
r_t
=
-\lambda_{\mathrm{step}}
+
R^{\mathrm{prog}}
\left(
\rho_{t+1}^m-\rho_t^m
\right)
+
\delta_t^m
\left(
R^\mathrm{goal}_{\mathrm{success}}y_t
-
R^\mathrm{goal}_{\mathrm{fail}}(1-y_t)
\right).
\label{eq:subgoal_reward}
\end{equation}
The first term is a small step penalty that discourages redundant actions. The second term assigns dense positive credit to transitions inside achieved key-state segments. The last term provides a terminal bonus when the assigned key state is achieved and a terminal penalty when the key state remains unfulfilled.
The reward-augmented offline dataset is written as: 
\begin{equation}
    \widetilde{\mathcal{D}}
=
\{(o_t,l,a_t,g_t,y_t,r_t)\}.
\end{equation}
With the annotated rewards, we train a value function \(V_\psi(o_t,l)\) to estimate the expected discounted return induced by the key-state relabeled rewards. For value learning, we use an \(n\)-step Bellman backup to construct the temporal-difference target:
\begin{equation}
\hat{V}_t
=
\sum_{i=0}^{n-1}
\gamma^i r_{t+i}
+
\gamma^n
V_{\bar{\psi}}(o_{t+n},l),
\label{eq:value_target}
\end{equation}
where \(V_{\bar{\psi}}\) denotes a target value network or a stop-gradient copy of the value model. In implementation, \(V_\psi\) is obtained by attaching a lightweight value head to the intermediate representation of the VLA backbone. 
The value function is optimized by minimizing the temporal-difference error:
\begin{equation}
\mathcal{L}_{V}(\psi)
=
\mathbb{E}_{(o_t,l)\sim\widetilde{\mathcal{D}}}
\left[
\left(
V_\psi(o_t,l)-\hat{V}_t
\right)^2
\right].
\label{eq:value_loss}
\end{equation}
Through this objective, states that are followed by achieved key-state progress are assigned higher values, while states in failed or incomplete segments receive lower values due to the absence of progress rewards and the terminal failure penalty.

\subsection{Offline Goal-conditioned Policy Optimization}
In this work, we seek a lightweight policy improvement objective that can be seamlessly integrated into the existing action modeling loss of a pretrained VLA policy. Concretely, we adopt an AWAC-style actor update~\cite{awac}, which casts policy improvement as advantage-weighted behavior regression. 

We calculate the advantage value for each action \(a_t\) using an \(n\)-step return estimate induced by the goal-conditioned rewards:
\begin{equation}
A_t
=
\left(
\sum_{i=0}^{n-1}
\gamma^i r_{t+i}
+
\gamma^n V_{\psi}(o_{t+n},l)
\right)
-
V_\psi(o_t,l).
\label{eq:offline_advantage}
\end{equation}
A positive advantage indicates that the observed action leads to a higher return than expected from the current state, while a small or negative advantage suggests lower contribution to future progress. 
Following AWAC~\cite{awac}, we convert the advantage into an exponential policy improvement weight:
\begin{equation}
\omega_t
=
\operatorname{sg}
\left[
\min
\left(
\exp\left(\frac{A_t}{\beta}\right),
\omega_{\max}
\right)
\right],
\label{eq:advantage_weight}
\end{equation}
where \(\beta\) controls the sharpness of advantage weighting, \(\omega_{\max}\) prevents a small number of high-advantage samples from dominating the update, and \(\operatorname{sg}(\cdot)\) denotes the stop-gradient operation.

We then optimize the policy with an AWAC-style offline RL loss over the entire relabeled dataset 
$\widetilde{\mathcal{D}}$: 
\begin{equation}
\mathcal{L}_{\mathrm{RL}}(\theta)
=
-
\mathbb{E}_{\tilde{x}_t\sim\widetilde{\mathcal{D}}}
\left[
\omega_t
\log \pi_\theta(a_t\mid o_t,l)
\right].
\label{eq:offline_rl_loss}
\end{equation}
This objective biases policy improvement toward high-advantage actions in the relabeled dataset:  actions that contribute to achieved key states receive larger weights, while actions with low progress return have limited influence on the policy update. As a result, the policy is encouraged not merely to imitate demonstrated actions, but to improve its ability to recover from intermediate states and continue progressing toward the intended goal states.

Furthermore, inspired by recent post-training paradigms in large language models (LLMs) that unify supervised and reinforcement objectives within a single optimization stage\cite{srft,uft,sftrldy}, we further retain SFT in the policy objective. Specifically, SFT is applied to successful demonstrations as a lightweight action prior, preserving low-level action feasibility and maintaining the language-action alignment inherited from the pretrained VLA model.
\begin{equation}
\mathcal{L}_{\mathrm{SFT}}(\theta)
=
-
\mathbb{E}_{x_t\sim\mathcal{D}_{\mathrm{demo}}}
 \left[
 \log \pi_\theta(a_t\mid o_t,l)
\right].
\label{eq:sft_loss}
\end{equation}
The final policy optimization objective is:
\begin{equation}
\mathcal{L}_{\mathrm{total}}(\theta)
=
\lambda_{\mathrm{RL}}\mathcal{L}_{\mathrm{RL}}(\theta)
+
\lambda_{\mathrm{SFT}}\mathcal{L}_{\mathrm{SFT}}(\theta)
\label{eq:total_loss}
\end{equation}

SFT and offline RL play complementary roles in this objective. SFT provides a coarse-grained behavioral prior, anchoring the policy to the feasible action distribution of expert demonstrations and stabilizing optimization. In contrast, offline RL provides fine-grained goal-directed optimization signals through goal-conditioned process rewards and advantage-weighted policy improvement, enabling the policy to deviate from demonstration-specific action paths when necessary and favor behaviors that better advance toward task-progress key states. 


For notational simplicity, we represent the action-matching term in the above objectives using the likelihood-style form \(\log \pi_\theta(a_t\mid o_t,l)\). This term serves as a general action-matching objective that encourages the policy output to align with the demonstrated action \(a_t\). In practice, it is instantiated according to the action head of the underlying VLA model, such as action-token likelihood~\cite{openvla}, continuous action regression~\cite{openvlaoft}, denoising prediction~\cite{diffusionpolicy,octo}, or flow-matching training objectives~\cite{pi0,pi05}.

\section{Experiments}

\subsection{Experimental Setup}
\paragraph{Benchmarks.}
To evaluate the effectiveness of our method, we conduct comprehensive experiments on two widely used robotic manipulation benchmarks, LIBERO~\cite{libero} and MetaWorld~\cite{metaworld}. To further assess robustness and performance under off-demonstration conditions, we design two additional evaluation settings. First, we use LIBERO-Plus~\cite{liberoplus} to test distribution shifts in camera viewpoint, robot initial state, visual noise, and object layout. Second, to evaluate dynamic recovery under execution-time disturbances, we introduce \textbf{LIBERO-ReAct}, a mid-execution perturbation benchmark built on LIBERO~\cite{libero}. LIBERO-ReAct includes action noise, gripper-state perturbations, and end-effector position perturbations, and is designed to evaluate goal-directed recoverability during policy execution.

\paragraph{Backbones and Baselines.}
OGPO is built on the $\pi_{0}$ and $\pi_{0.5}$ backbones. We compare with their SFT-based few-shot and full-shot post-training variants to evaluate improvements over standard supervised imitation. We further report representative VLA models, including OpenVLA~\cite{openvla}, SmolVLA~\cite{smolvla}, UniVLA~\cite{univla}, and OpenVLA-OFT~\cite{openvlaoft}, as broader references, and include PPO-based online RL baselines from RLinf~\cite{pirl} for comparison with online RL post-training. Overall, this baseline set provides a comprehensive and rigorous evaluation for assessing the effectiveness of our OGPO.


\subsection{Main Results}

\begin{table*}[t]
\centering
\caption{
Performance comparison on the LIBERO benchmark. Policy input notation: T (third-view image), I (instruction), W (wrist image), P (proprioception). OGPO is built upon the few-shot SFT policy, with $\uparrow$ indicating performance gains over the few-shot baseline.
}
\label{tab:libero_main}
\begingroup
\resulttabfont
\resulttabstretch
\setlength{\tabcolsep}{4.5pt}
\begin{tabular}{@{}l c c c c c c@{}}
\toprule
\multirow{2}{*}{\textbf{Model}} 
& \multirow{2}{*}{\textbf{Input}} 
& \multicolumn{5}{c}{\textbf{LIBERO}} \\
\cmidrule(lr){3-7}
& & \textbf{Spatial} & \textbf{Object} & \textbf{Goal} & \textbf{Long} & \textbf{Avg.} \\
\midrule
OpenVLA\cite{openvla} & T+I & 84.7 & 88.4 & 79.2 & 53.7 & 76.5 \\
$\pi_0$-fast\cite{pi0fast}  & T+W+P+I & 96.4 & 96.8 & 88.6 & 60.2 & 85.5 \\
$\pi_0$\cite{pi0} & T+W+P+I & 96.8 & 98.8 & 95.8 & 85.2 & 94.2 \\
SmolVLA\cite{smolvla} & T+W+P+I & 93.0 & 94.0 & 91.0 & 77.0 & 88.8 \\
WorldVLA\cite{worldvla} & T+I & 85.6 & 89.0 & 82.6 & 59.0 & 79.1 \\
NORA\cite{nora} & T+I & 92.2 & 95.4 & 89.4 & 74.6 & 87.9 \\
CoT-VLA\cite{cotvla} & T+I & 87.5 & 91.6 & 87.6 & 69.0 & 81.1 \\
UniVLA\cite{univla} & T+I & 96.5 & 96.8 & 95.6 & 92.0 & 95.2 \\
MolmoAct\cite{molmoact}  & T+I & 87.0 & 95.4 & 87.6 & 77.2 & 86.6 \\
ThinkAct\cite{thinkact} & T+I & 88.3 & 91.4 & 87.1 & 70.9 & 84.4 \\
GROOT N1\cite{groot}  & T+I & 94.4 & 97.6 & 93.0 & 90.6 & 93.9 \\
OpenVLA-OFT\cite{openvlaoft} & T+W+P+I & 96.2 & 98.3 & 96.2 & 90.7 & 95.3 \\
$\pi_{0.5}$-Full\cite{pi05} & T+W+P+I & 98.8 & 98.2 & 98.0 & 92.4 & 96.85 \\
\midrule
TGRPO\cite{tgrpo} & T+I & 90.4 & 92.2 & 81.0 & 59.2 & 80.7 \\
VLA-RL\cite{vla-rl} & T+I & 90.2 & 91.8 & 82.2 & 59.8 & 81.0 \\
SimpleVLA-RL\cite{simplevlarl} & T+I & 98.2 & 98.7 & 98.8 & 91.7 & 96.9 \\
$\pi_{0.5}$-PPO (RLinf)\cite{pirl} & T+W+P+I & 99.4 & 99.8 & 98.8 & 94.0 & 98.0 \\
\midrule
$\pi_{0.5}$-Few Shot & T+W+P+I & 84.6 & 95.4 & 84.6 & 43.9 & 77.1 \\
\rowcolor{blue!12}
+ OGPO & T+W+P+I & 99.2 & 98.6 & 97.6 & 93.2 & 97.1 \\
 &  & \textcolor{green!60!black}{$\uparrow$14.6} 
    & \textcolor{green!60!black}{$\uparrow$3.2} 
    & \textcolor{green!60!black}{$\uparrow$13.0} 
    & \textcolor{green!60!black}{$\uparrow$49.7} 
    & \textcolor{green!60!black}{$\uparrow$20.0} \\
\bottomrule
\end{tabular}
\endgroup
\end{table*}

Our evaluation begins with the overall performance on widely used manipulation benchmarks. 
Tables~\ref{tab:libero_main} and~\ref{tab:metaworld_main} present the  results on LIBERO and MetaWorld-MT50. 
On LIBERO, OGPO and PPO are initialized from the same $\pi_{0.5}$ few-shot policy, obtained with limited downstream demonstrations. Starting from this backbone, OGPO improves the average success rate from 77.1\% to 97.1\%, outperforming the $\pi_{0.5}$ full-SFT baseline of 96.85\% and achieving performance comparable to $\pi_{0.5}$+PPO, an online RL method requiring costly online rollout. The improvement is particularly significant on the long-horizon \textbf{LIBERO-Long} suite, where our method raises the success rate from 43.9\% to 93.2\%, also surpassing the full-SFT result of 92.4\%, indicating its enhanced goal-reaching capability in long-horizon manipulation tasks.
On MetaWorld-MT50, our method consistently outperforms SFT across different backbones, improving $\pi_0$ from 50.8\% to 81.0\% and approaching the online PPO-based result of 85.8\%. The gains are especially large on more challenging tasks; for example, on the $\pi_0$ backbone, our method improves the success rate by 28.4\% on Hard tasks and 53.6\% on \textbf{Very Hard} tasks. Overall, these results validate that OGPO can effectively extract goal-reaching supervision from offline trajectories, enabling substantial policy improvement without online rollouts, particularly on long-horizon and difficult manipulation tasks.

\begin{table*}[t]
\centering
\caption{
Performance comparison on the MetaWorld-MT50 benchmark. Policy input notation: T (third-view image), P (proprioception), C (task conditioning). C denotes the task information used to distinguish MetaWorld tasks, and may be instantiated as a language instruction, task ID, or task embedding depending on the model. 
}
\label{tab:metaworld_main}
\begingroup
\resulttabfont
\resulttabstretch
\setlength{\tabcolsep}{6pt}
\begin{tabular}{@{}l c c c c c c@{}}
\toprule
\multirow{2}{*}{\textbf{Model}} 
& \multirow{2}{*}{\textbf{Input}} 
& \multicolumn{5}{c}{\textbf{MetaWorld-MT50}} \\
\cmidrule(lr){3-7}
& & \textbf{Easy} & \textbf{Medium} & \textbf{Hard} & \textbf{Very Hard} & \textbf{Avg.} \\
\midrule
Diffusion Policy\cite{diffusionpolicy} & T+C & 23.1 & 10.7 & 1.9 & 6.1 & 10.5 \\
TinyVLA\cite{tinyvla} & T+P+C & 77.6 & 21.5 & 11.4 & 15.8 & 31.6 \\
SmolVLA\cite{smolvla} & T+P+C & 87.1 & 51.8 & 70.0 & 64.0 & 68.2 \\
$\pi_0$-PPO (RLinf)\cite{pirl} & T+P+I & 91.1 & 81.8 & 78.3 & 92.0 & 85.8 \\
\midrule
$\pi_0$-SFT & T+P+C & 77.9 & 51.8 & 53.3 & 20.0 & 50.8 \\
\rowcolor{blue!12}
$\pi_0$-OGPO & T+P+C & 89.2 & 79.6 & 81.7 & 73.6 & 81.0 \\
 &  & \textcolor{green!60!black}{$\uparrow$11.3} 
    & \textcolor{green!60!black}{$\uparrow$27.8} 
    & \textcolor{green!60!black}{$\uparrow$28.4} 
    & \textcolor{green!60!black}{$\uparrow$53.6} 
    & \textcolor{green!60!black}{$\uparrow$30.2} \\
\midrule
$\pi_{0.5}$-SFT & T+P+C & 68.2 & 37.3 & 41.7 & 28.0 & 43.8 \\
\rowcolor{blue!12}
$\pi_{0.5}$-OGPO & T+P+C & 82.1 & 55.8 & 71.7 & 59.0 & 67.2 \\
 &  & \textcolor{green!60!black}{$\uparrow$13.9} 
    & \textcolor{green!60!black}{$\uparrow$18.5} 
    & \textcolor{green!60!black}{$\uparrow$30.0} 
    & \textcolor{green!60!black}{$\uparrow$31.0} 
    & \textcolor{green!60!black}{$\uparrow$23.4} \\
\bottomrule
\end{tabular}
\endgroup
\end{table*}

Furthermore, to evaluate the robustness of OGPO and its ability to handle off-demonstration deviations, Table~\ref{tab:libero_robustness} reports performance under selected LIBERO-Plus perturbations and our proposed \textbf{LIBERO-ReAct}, a mid-execution perturbation benchmark built on LIBERO~\cite{libero}. Among the perturbations provided by LIBERO-Plus, we select four representative types: camera viewpoint changes, robot initial-state variations, visual noise, and object layout changes, covering scene-, observation-, and initial-state distribution shifts. To further test dynamic recovery during policy execution, LIBERO-ReAct introduces three mid-episode action-space perturbations across the four LIBERO task suites, including Spatial, Object, Goal, and Long. For each task, we randomly select a 20-step interval during the episode and apply the perturbation after the policy predicts an action and before the action is executed in the environment: 
\begin{itemize}
    \item \textbf{Action noise}: we add uniform noise to the executed action command, with a range of $[-0.1, 0.1]$ for arm control and $[-0.5, 0.5]$ for gripper control.
    \item \textbf{Gripper-state perturbation}: we flip the gripper command to the opposite state, forcing an open command to close and a close command to open.
    \item \textbf{End-effector position perturbation}: we add uniform noise within $[-0.2, 0.2]$ to the predicted position motion command, simulating spatial deviations of the end effector.
\end{itemize}
Compared with the full-shot SFT baseline, our offline fine-tuned $\pi_{0.5}$ improves the overall robustness score from 76.7\% to 81.8\%. It achieves clear gains on LIBERO-Plus Robot-Init and Noise perturbations, and shows even larger improvements under LIBERO-ReAct disturbances. In particular, the success rate under gripper-state perturbation increases from 76.8\% to 87.7\%, suggesting that our method better recovers from erroneous grasping or releasing actions. These results indicate that our method improves both robustness to static distribution shifts and recovery from execution-time disturbances.

\begin{table*}[t]
\centering
\caption{
Robustness Evaluation on the LIBERO-Plus Benchmark and Our Extended Perturbation Settings. The red-highlighted columns indicate the additional mid-episode perturbations introduced by us. All results are evaluated without perturbation-specific fine-tuning.
}
\label{tab:libero_robustness}
\resizebox{\textwidth}{!}{
\begin{tabular}{lcccccccc}
\toprule
\multirow{2}{*}{\textbf{Model}} 
& \multicolumn{7}{c}{\textbf{Perturbation Dimensions}} 
& \multirow{2}{*}{\textbf{Total}} \\
\cmidrule(lr){2-8}
& \textbf{Camera} 
& \textbf{Robot-Init}
& \textbf{Noise} 
& \textbf{Layout} 
& \textcolor{red}{\textbf{Action}}
& \textcolor{red}{\textbf{Gripper}}
& \textcolor{red}{\textbf{End-effector}}

&  \\
\midrule
$\pi_0$\cite{pi0}              & 13.8 & 6.0  & 79.0 & 68.9 & 81.5 & 9.3 & 62.4 & 45.8 \\
$\pi_0$+fast\cite{pi0fast}         & 65.1 & 21.6 & 74.4 & 68.8 & 75.2 & 19.8 & 63.8 & 55.5 \\
OpenVLA\cite{openvla}          & 0.8  & 3.5  & 15.2 & 28.5 & 19.7 & 24.3 & 31.5  & 17.6\\
OpenVLA-OFT\cite{openvlaoft}      & 56.4 & 31.9 & 75.8 & 74.2 & 85.2 & 84.4 & 87.6 & 70.8 \\
\midrule
$\pi_{0.5}$-SFT\cite{pi05}    & 62.8 & 69.8 & 69.5  & 85.4 & 86.6 & 76.8 & 86.0 & 76.7 \\
\rowcolor{blue!12}
$\pi_{0.5}$-OGPO     & 61.5 & 76.6 & 73.1  & 86.6 & 96.2 & 87.7 & 90.1 & 81.8 \\
                  & \textcolor{green!60!black}{$\downarrow$1.3} 
                  & \textcolor{green!60!black}{$\uparrow$6.8} 
                  & \textcolor{green!60!black}{$\uparrow$3.6} 
                  & \textcolor{green!60!black}{$\uparrow$1.2} 
                  & \textcolor{green!60!black}{$\uparrow$9.6} 
                  & \textcolor{green!60!black}{$\uparrow$10.9} 
                  & \textcolor{green!60!black}{$\uparrow$4.1} 
                  & \textcolor{green!60!black}{$\uparrow$5.1} \\

\bottomrule
\end{tabular}
}
\end{table*}

\begin{figure}[t]
    \centering

    \begin{subfigure}[t]{0.38\linewidth}
        \centering
        \includegraphics[width=\linewidth]{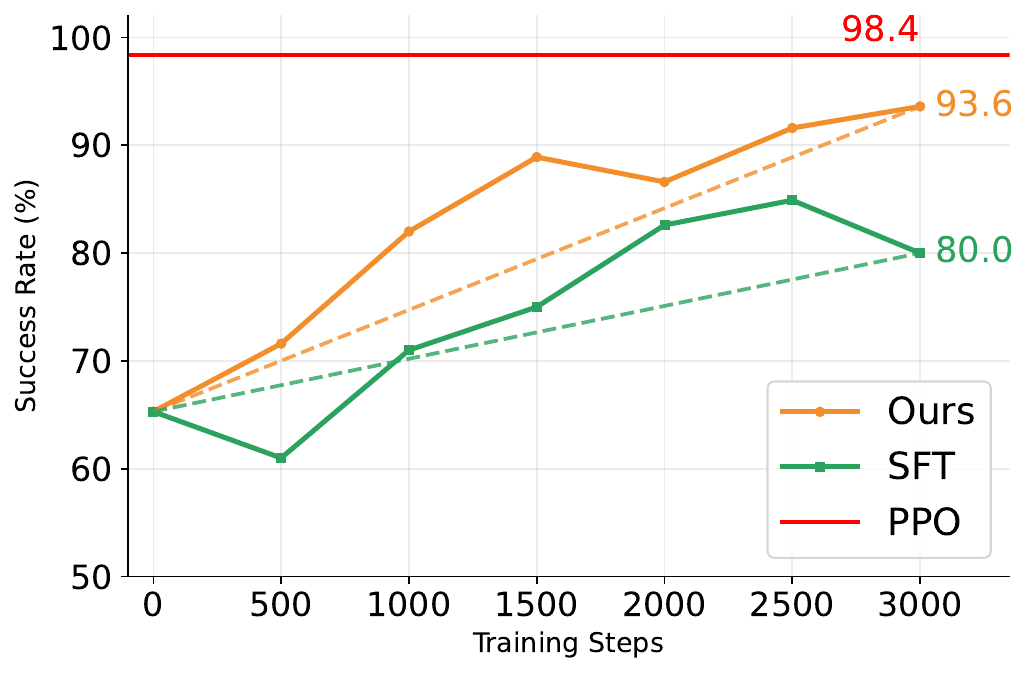}
        \caption{LIBERO-Spatial}
        \label{fig:libero_spatial_t}
    \end{subfigure}
    \hspace{0.02\linewidth}
    \begin{subfigure}[t]{0.38\linewidth}
        \centering
        \includegraphics[width=\linewidth]{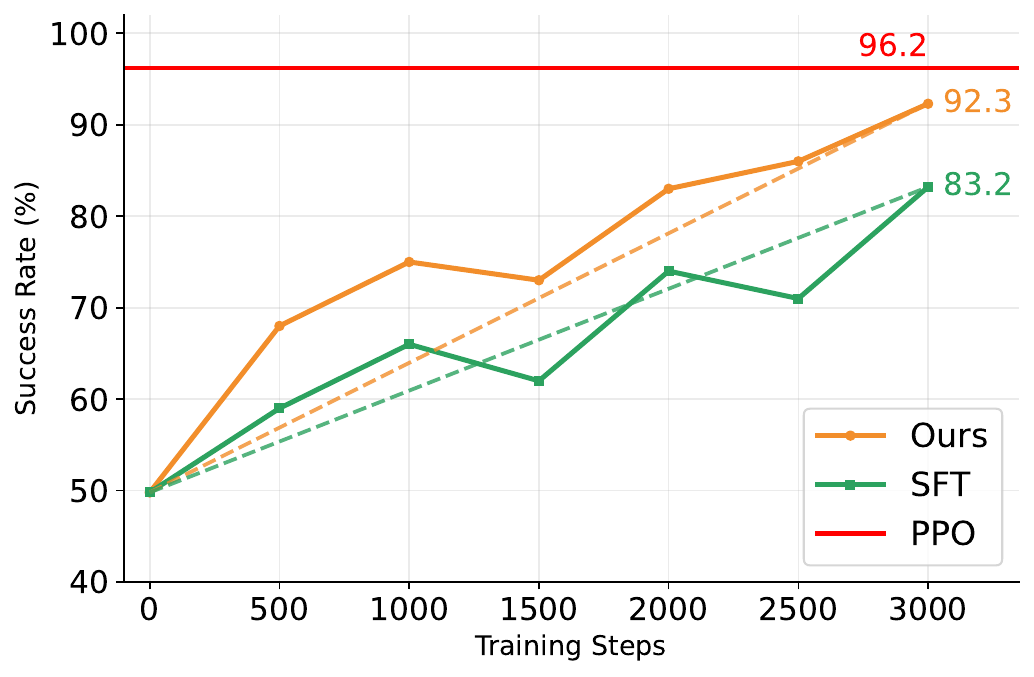}
        \caption{LIBERO-Goal}
        \label{fig:libero_goal_t}
    \end{subfigure}

    \caption{Training efficiency comparison between Ours and SFT using the $\pi_0$ backbone.}
    \label{fig:training_efficiency}
\end{figure}

At last, to evaluate the scalability of OGPO, we conduct an additional set of experiments using $\pi_0$ as the backbone. As shown in Fig.~\ref{fig:training_efficiency}, OGPO also achieves substantial improvements over standard SFT on LIBERO-Spatial and LIBERO-Goal under the same 3,000 training steps. Specifically, our method reaches 93.6\% on LIBERO-Spatial and 92.3\% on LIBERO-Goal, outperforming SFT by 13.6 and 11.1 points, respectively. These results further validate the effectiveness of OGPO, suggesting that the proposed offline objective provides more effective learning signals than pure supervised imitation and leads to more efficient policy improvement.

\subsection{Ablation Study}

To better understand the contribution of each component, we conduct ablation studies from two perspectives. First, we examine the reward design, which provides the basis for offline advantage estimation. Second, we analyze the policy optimization objective by removing each training signal, including offline RL and SFT. 

\begin{figure}[t]
    \centering

    \begin{subfigure}[t]{0.48\linewidth}
        \centering
        \includegraphics[width=\linewidth]{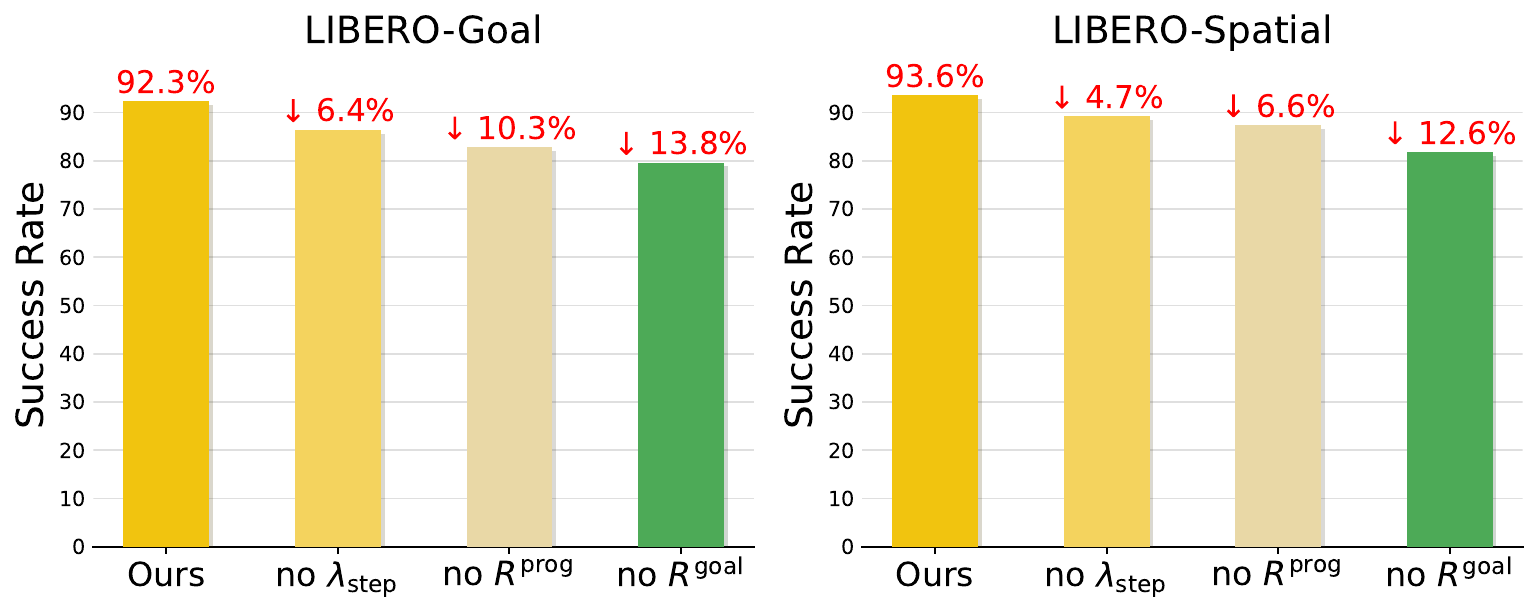}
        \caption{Ablation on reward designs.}
        \label{fig:libero_spatial_r}
    \end{subfigure}
    \hspace{0.01\linewidth}
    \begin{subfigure}[t]{0.48\linewidth}
        \centering
        \includegraphics[width=\linewidth]{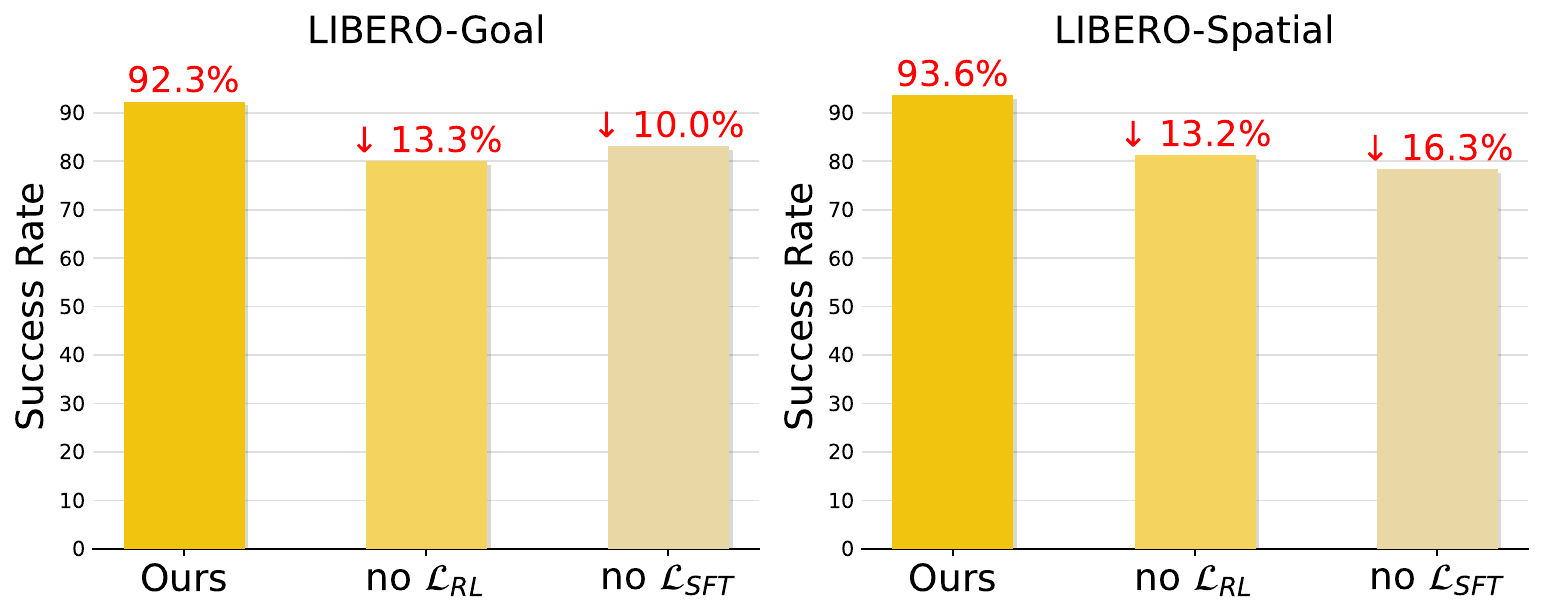}
        \caption{Ablation on policy fine-tuning objectives.}
        \label{fig:libero_goal_r}
    \end{subfigure}

    \caption{Ablation studies using the $\pi_0$ backbone trained for 3,000 steps on LIBERO-Goal and LIBERO-Spatial. }
    \label{fig:abla}
\end{figure}

\paragraph{Ablation on reward designs.}
As shown in Fig.~\ref{fig:abla}(a), the proposed reward design achieves the best performance on both LIBERO-Goal and LIBERO-Spatial. Among all variants, removing \(R^{\mathrm{goal}}\) leads to the largest performance drop. This indicates that subgoal-level completion signals are crucial for offline value learning: without trajectory segmentation and subgoal relabeling, the value function cannot effectively distinguish achieved progress from failed segments.

\paragraph{Ablation on policy fine-tuning objectives.}
As shown in Fig.~\ref{fig:abla}(b), the full objective performs best across both tasks. Removing the offline RL term clearly degrades performance, highlighting the importance of AWAC-style advantage-weighted policy improvement for learning higher-return behaviors. Meanwhile, the decline observed after removing SFT confirms its role as positive behavior regularization, which preserves feasible successful behaviors and prevents the AWAC-style update from drifting away from the pretrained action prior.

\section{Conclusion and Limitations}
In this work, we investigate how to leverage offline data to train a robust VLA model capable of recovering toward goal states after deviating from demonstration trajectories. Specifically, we propose OGPO, an offline goal-conditioned policy optimization framework for recoverable VLA post-training. Instead of treating demonstrations as step-wise action supervision, OGPO relabels offline trajectories into key-state-centered goal-reaching segments and transforms segment completion signals into goal-conditioned process rewards. These rewards facilitate value learning and advantage-weighted policy optimization. In addition, OGPO retains SFT to provide a coarse-grained behavioral prior that anchors the policy to the feasible action distribution induced by expert demonstrations. Overall, OGPO shifts VLA post-training from trajectory-level action imitation toward key-state reachability learning, thereby improving goal-directed recoverability without requiring online interaction. Extensive experiments demonstrate the effectiveness, robustness, and recovery capability of OGPO under off-demonstration scenarios. Despite the promising results, this work still has several limitations. First, the current gripper-state-based segmentation strategy may not generalize well to tasks whose progress cannot be naturally characterized by such events. Second, as a purely offline framework, OGPO is  limited by the quality of the offline dataset, which may restrict its ability to recover from severe out-of-distribution deviations beyond the demonstrated trajectories. In future work, we plan to investigate more general and robust goal-state relabeling strategies, along with data generation and augmentation methods that can improve the diversity of offline data for recoverable policy learning.


\bibliographystyle{plain}   
\bibliography{main}        

\medskip

\newpage

\appendix

\section{Pseudo-code of OGPO}
\label{app:ogpo_algorithm}

Algorithm~\ref{alg:ogpo} summarizes the training procedure of \textsc{OGPO}. 
The relabeled goals are only used to construct process rewards during training, and are not introduced as additional inputs to the policy or value function.

\begin{algorithm}[H]
\caption{\textsc{OGPO}: Offline Goal-conditioned Policy Optimization}
\label{alg:ogpo}
\begin{algorithmic}[1]
\Require Successful demonstrations $\mathcal{D}_{\mathrm{demo}}$, failed trajectories $\mathcal{D}_{\mathrm{fail}}$, pretrained VLA policy $\pi_\theta$, value function $V_\psi$, VLM annotator
\Ensure Optimized policy $\pi_\theta$

\State Initialize relabeled dataset $\widetilde{\mathcal{D}} \leftarrow \emptyset$

\For{each trajectory $\tau \in \mathcal{D}_{\mathrm{demo}} \cup \mathcal{D}_{\mathrm{fail}}$}
    \State Segment $\tau$ into stages $\{\tau^1,\ldots,\tau^M\}$ using gripper-state changes
    \For{each segment $\tau^m$}
        \If{$\tau \in \mathcal{D}_{\mathrm{demo}}$}
            \State Set the segment terminal state as goal $g^m$ and label $y^m \leftarrow 1$
        \Else
            \State Use the VLM annotator to infer goal $g^m$ and completion label $y^m$
        \EndIf
        \State Add goal labels to transitions in $\tau^m$
    \EndFor
\EndFor

\For{each relabeled segment $\tau^m$}
    \State Compute progress variable $\rho_t^m$ using Eq.~\ref{eq:progress_variable}
    \State Annotate process rewards $r_t$ using Eq.~\ref{eq:subgoal_reward}
    \State Add reward-augmented transitions to $\widetilde{\mathcal{D}}$
\EndFor

\State \textbf{Value learning:} Train $V_\psi$ on $\widetilde{\mathcal{D}}$ with the $n$-step target in Eq.~\ref{eq:value_target} by minimizing $\mathcal{L}_{V}$ in Eq.~\ref{eq:value_loss}

\State \textbf{Policy optimization:}
\While{not converged}
    \State Sample a relabeled minibatch from $\widetilde{\mathcal{D}}$ 
    \State Compute advantages $A_t$ and AWAC weights $\omega_t$ using the trained value function $V_\psi$
    \State Compute $\mathcal{L}_{\mathrm{RL}}$ in Eq.~\ref{eq:offline_rl_loss} and $\mathcal{L}_{\mathrm{SFT}}$ in Eq.~\ref{eq:sft_loss}
    \State Update $\pi_\theta$ by minimizing $\mathcal{L}_{\mathrm{total}}$ in Eq.~\ref{eq:total_loss}
\EndWhile

\State \Return $\pi_\theta$
\end{algorithmic}
\end{algorithm}

\section{Detailed Trajectory Segmentation and Goal Relabeling}
\label{app:segmentation}

This section provides the implementation details of our trajectory segmentation and goal relabeling procedure. 
The key idea is to exploit the gripper open-close pattern in manipulation trajectories. 
In many tabletop manipulation tasks, gripper state changes naturally correspond to important semantic transitions, such as approaching an object, grasping it, transporting it, and releasing it at the target location. 
Therefore, we use gripper state changes as temporal anchors to divide a long trajectory into several goal-level segments. 
After segmentation, each segment is assigned a semantic goal label and a binary completion label.

\subsection{Gripper-Based Trajectory Segmentation}
\label{app:gripper_segmentation}

Given an offline trajectory
\[
\tau = \{(o_t,l,a_t)\}_{t=1}^{T},
\]
where \(o_t\) is the observation, \(l\) is the language instruction, and \(a_t\) is the action, we extract the gripper command from the action sequence:
\[
c_t \in \{\mathrm{open}, \mathrm{close}\}.
\]
We then identify gripper switching points as candidate segment boundaries:
\[
\mathcal{B}
=
\{t \mid c_t \neq c_{t-1}\}.
\]
To avoid noisy boundaries caused by short-term command oscillations, we only keep a switching point if the new gripper state is maintained for a short temporal window. 
The final boundary set is formed by adding the beginning and the end of the trajectory:
\[
\bar{\mathcal{B}} = \{1\} \cup \mathcal{B} \cup \{T\}.
\]
The trajectory is then divided into a sequence of segments:
\[
\tau = \{\tau^1,\tau^2,\ldots,\tau^M\},
\qquad
\tau^m = \{(o_t,l,a_t)\}_{t=s_m}^{e_m},
\]
where \(s_m\) and \(e_m\) are two adjacent boundaries in \(\bar{\mathcal{B}}\).

This gripper-based segmentation is simple but effective for manipulation tasks because most meaningful object interactions are synchronized with gripper events. 
For example, an open-to-close transition near an object usually indicates a grasping attempt, while a close-to-open transition near a target region usually indicates a placing or releasing attempt. 
The segment between these two events often corresponds to a transportation or alignment goal.

\subsection{Goal Relabel}
\label{app:goal_assignment}

After obtaining the temporal segments, we assign a semantic goal \(g^m\) to each segment \(\tau^m\). 
The goal label is determined by the task instruction, the gripper transition type, and the visual state change within the segment. 
For example, for a task instruction such as ``put the black bowl in the bottom drawer of the cabinet and close it'', the trajectory can be decomposed into the following goals:
\[
\text{reach }
\rightarrow
\text{grasp}
\rightarrow
\text{move}
\rightarrow
\text{lay}
\rightarrow
\text{close}.
\]

In practice, we use the following rules to assign goal labels:
\begin{itemize}
    \item If the segment ends with an open-to-close gripper transition near a target object, it is labeled as a grasping-related goal.
    \item If the gripper remains closed and the object pose changes with the end-effector, the segment is labeled as a moving or transporting goal.
    \item If the segment ends with a close-to-open gripper transition near the target region, it is labeled as a placing or releasing goal.
    \item For articulated-object tasks, if the gripper remains closed while the joint state or object part pose changes, the segment is labeled as an opening, pulling, pushing, or closing goal.
\end{itemize}

This procedure converts each original trajectory into a sequence of goal-conditioned segments:
\[
\tau
\rightarrow
\{(\tau^1,g^1),(\tau^2,g^2),\ldots,(\tau^M,g^M)\}.
\]

\subsection{Completion Labeling}
\label{app:completion_labeling}

For each segment \(\tau^m\), we further assign a binary completion label:
\[
y^m \in \{0,1\},
\]
where \(y^m=1\) indicates that the segment successfully completes its assigned goal \(g^m\), and \(y^m=0\) indicates that the segment fails to complete the goal.

\paragraph{Successful trajectories.}
For a trajectory that successfully completes the full task, we label all its valid goal segments as successful:
\[
y^m = 1,\qquad \forall m.
\]

\paragraph{Failed trajectories.}
To better localize the failure stage and provide more accurate supervision for value learning, we use a VLM-based offline annotator to determine whether each segment in a failed trajectory completes its assigned goal. The VLM is asked to judge whether the visual state at the end of the segment satisfies the assigned goal. The resulting relabeled dataset is:
\[
\widetilde{\mathcal{D}}
=
\{(o_t,l,a_t,g_t,y_t)\},
\]
where \(g_t\) and \(y_t\) are inherited from the segment containing time step \(t\).

\subsection{VLM Annotation Prompt for Failed Segments}
\label{app:vlm_prompt}

For failed trajectories, we use Qwen2.5-VL-72B-Instruct as the VLM annotator to evaluate whether each segment completes its candidate goal. As shown in Figure~\ref{fig:vlm_annotation_prompt}, the prompt provides the task instruction, candidate goal, and representative segment observations, and asks the annotator to output a completion label with a short explanation.

\begin{figure}[t]
\centering
\begin{tcolorbox}[
    width=0.92\linewidth,
    colback=blue!6,
    colframe=black,
    colbacktitle=black,
    coltitle=white,
    title=\textbf{VLM Annotation Prompt},
    fonttitle=\normalsize,
    boxrule=0.8pt,
    arc=3pt,
    left=8pt,right=8pt,top=8pt,bottom=8pt,
    enhanced,
    attach boxed title to top left={xshift=10pt,yshift=-2pt}
]
\small
Your task is to determine whether a trajectory segment completes a given candidate subgoal.

\textbf{Task instruction:} \textcolor{red}{\{instruction\}}

\textbf{Candidate goal:} \textcolor{red}{\{goal\}}

\textbf{Segment observation:} The following images show the first, middle, and last frames of a trajectory segment.

\textbf{Question:} Does this segment successfully complete the candidate goal?

Please answer with one of the following labels:

\texttt{SUCCESS}: the final frame satisfies the candidate goal.

\texttt{FAILURE}: the final frame does not satisfy the candidate goal.

Also provide a short reason.
\end{tcolorbox}
\caption{Prompt template for VLM-based annotation of failed trajectory segments. We use Qwen2.5-VL-72B-Instruct as the VLM annotator.}
\label{fig:vlm_annotation_prompt}
\end{figure}

\subsection{Annotated Cases}
\label{app:annotated_cases}

\begin{figure*}[t]
    \centering
    \includegraphics[width=0.95\textwidth]{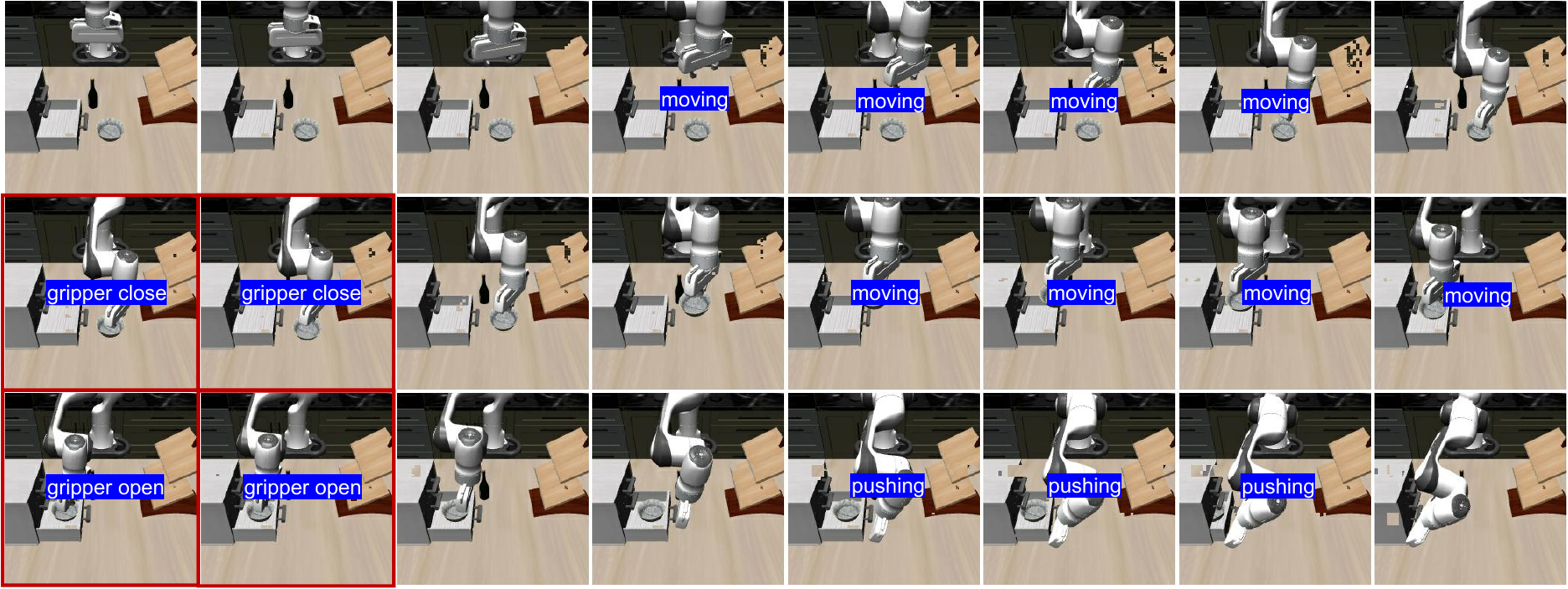}
    \caption{
    Annotated example of a successful trajectory. 
    }
    \label{fig:segmentation_case_success}
\end{figure*}

\begin{figure*}[t]
    \centering
    \includegraphics[width=0.95\textwidth]{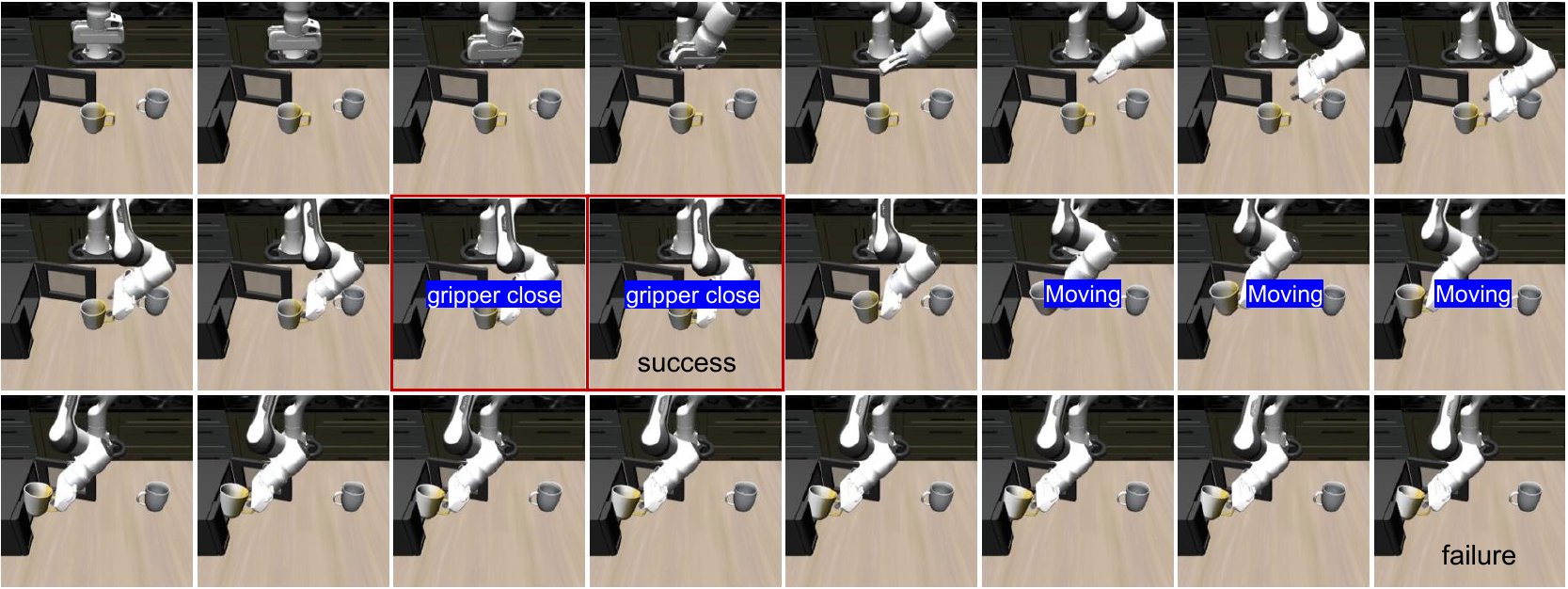}
    \caption{
    Annotated example of a failed trajectory. 
    }
    \label{fig:segmentation_case_failure}
\end{figure*}

Figure~\ref{fig:segmentation_case_success} and Figure~\ref{fig:segmentation_case_failure} show two annotated examples. For visualization clarity, we only show frames sampled every 10 time steps as a coarse illustration.
The first case is a successful trajectory. 
After gripper-based segmentation, all goal segments are labeled as achieved. 
The second case is a failed trajectory. 
Although the full trajectory fails, the early segments still complete meaningful goals, while the later segment fails to complete the assigned goal. 
Our VLM-based annotation preserves the achieved segments and marks the failed segment as negative experience.

\section{Reward Design and Value Analysis}
\label{app:reward_analysis}

This section provides additional details on the reward annotation procedure and analyzes the learned value estimates. 
Our reward design converts sparse goal completion labels into dense goal-conditioned learning signals. 
For a segment \(\tau^m\), we use the normalized progress variable \(\rho_t^m\) to assign dense progress rewards to completed segments, while failed segments do not receive progressive positive rewards. 
The reward consists of three parts: a step penalty, a dense progress reward, and a segment-end success or failure term.

\paragraph{Reward annotation.}
For each transition within a segment, the reward is computed according to Eq.~\ref{eq:subgoal_reward}. 
The step penalty discourages unnecessarily long or redundant action sequences. 
The progress reward provides smooth credit assignment for actions that move the system toward a completed goal. 
The terminal success bonus encourages completion of the assigned goal, while the terminal failure penalty prevents failed segments from being treated as useful demonstrations. In all experiments, we set 
\(\lambda_{\mathrm{step}} = 0.01\), 
\(R^{\mathrm{prog}} = 1.0\), 
\(R^\mathrm{goal}_{\mathrm{success}} = 1.0\), and 
\(R^\mathrm{goal}_{\mathrm{fail}} = 2.0\), 
unless otherwise specified.

\begin{figure}[t]
    \centering

    \begin{subfigure}{0.48\linewidth}
        \centering
        \includegraphics[width=\linewidth]{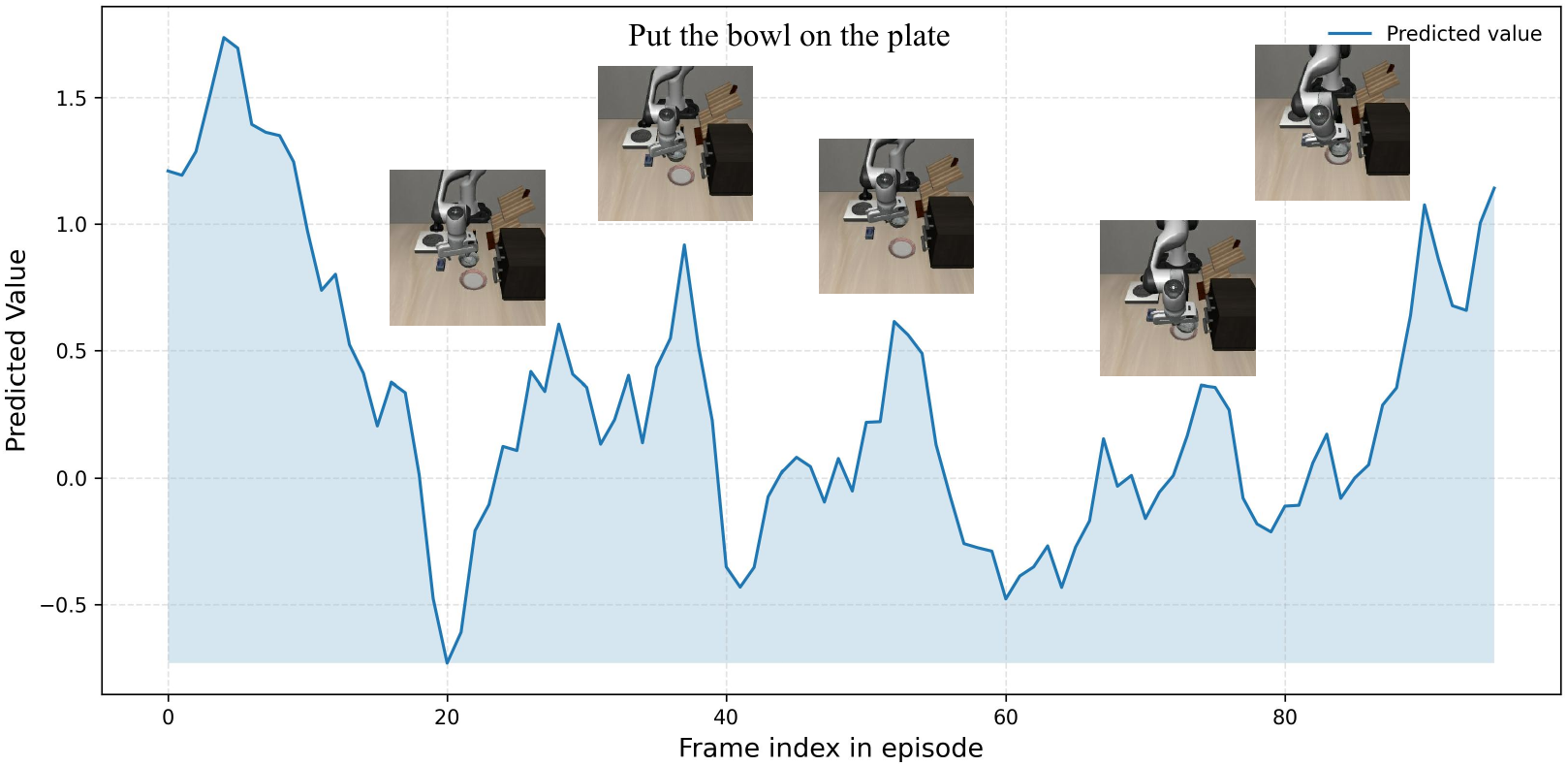}
        \caption{Put the bowl on the plate. }
        \label{fig:case1}
    \end{subfigure}
    \hfill
    \begin{subfigure}{0.48\linewidth}
        \centering
        \includegraphics[width=\linewidth]{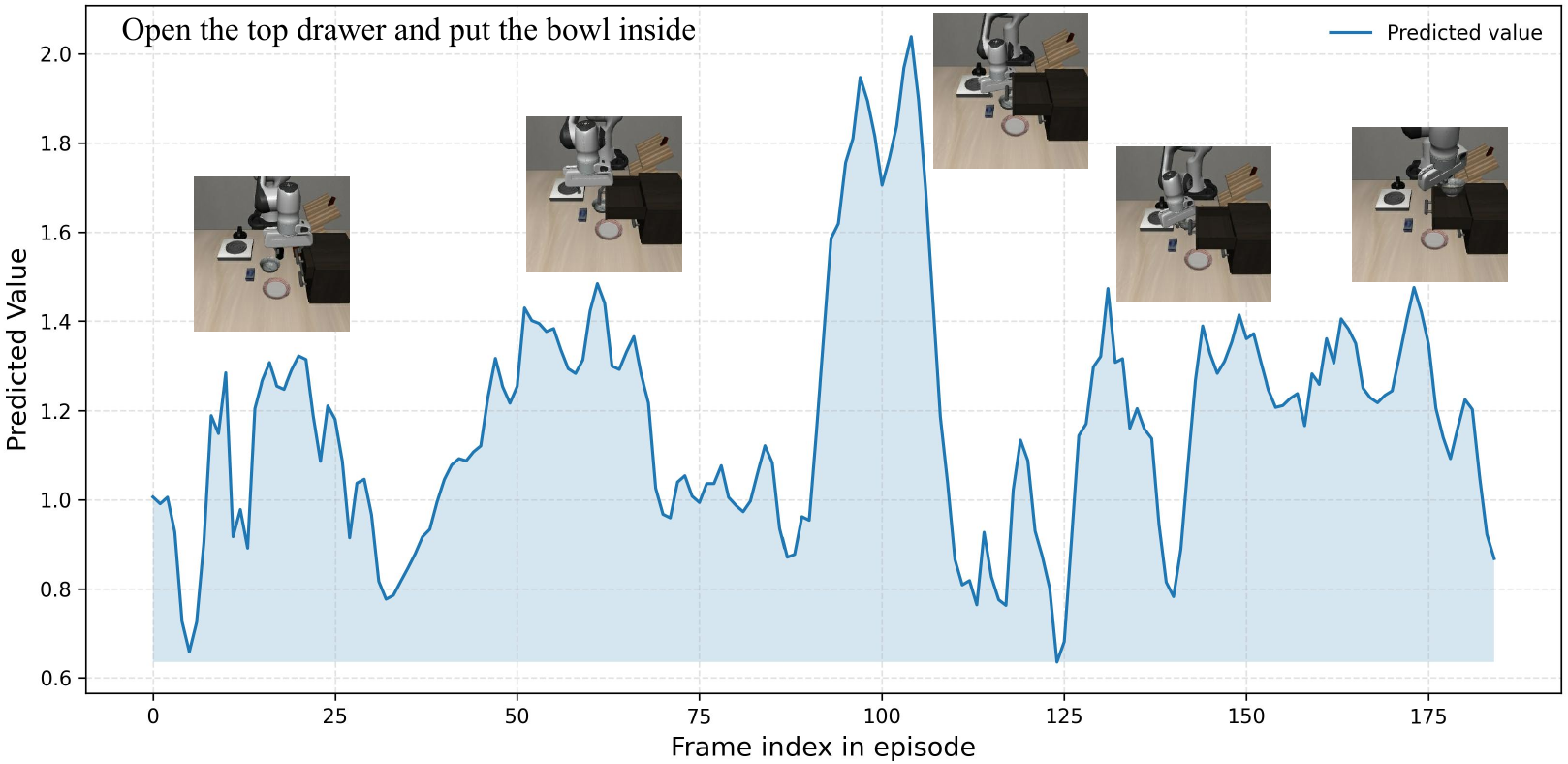}
        \caption{Open the drawer and put the bowl inside.}
        \label{fig:case2}
    \end{subfigure}

    \vspace{0.6em}

    \begin{subfigure}{0.48\linewidth}
        \centering
        \includegraphics[width=\linewidth]{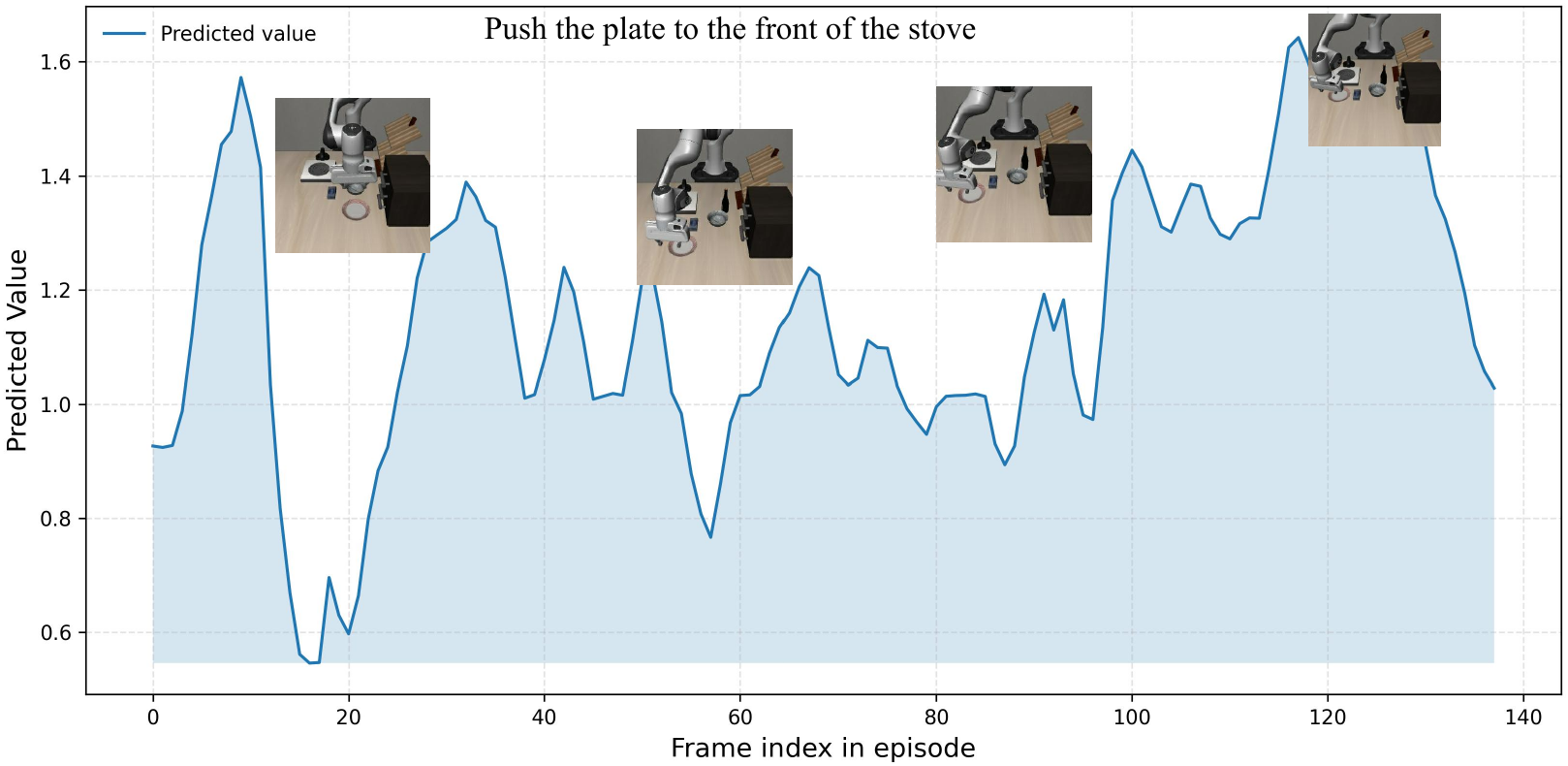}
        \caption{Push the plate to the front of the stove.}
        \label{fig:case3}
    \end{subfigure}
    \hfill
    \begin{subfigure}{0.48\linewidth}
        \centering
        \includegraphics[width=\linewidth]{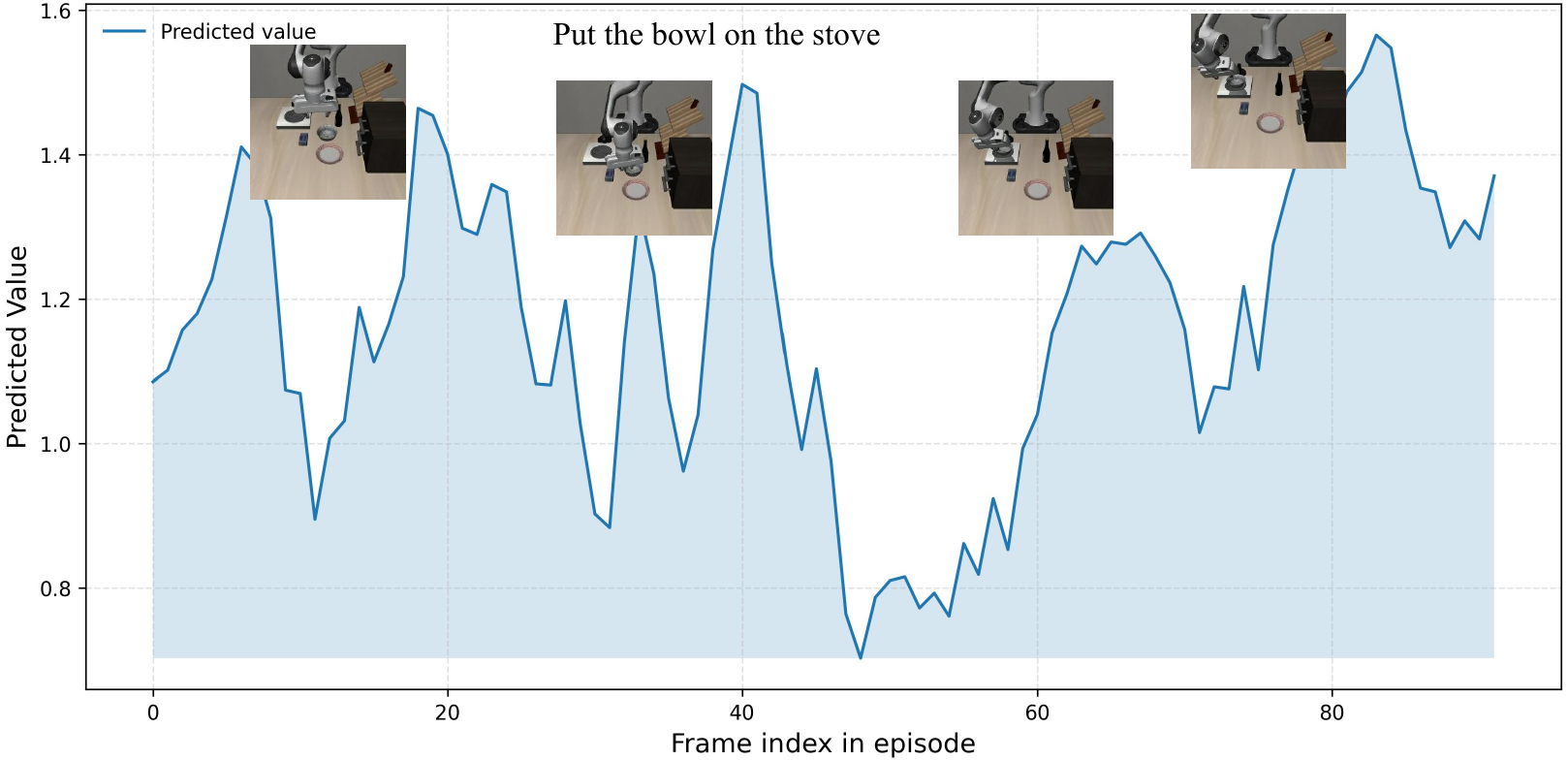}
        \caption{Put the bowl on the stove.}
        \label{fig:case4}
    \end{subfigure}

    \caption{
    Visualization of the learned goal-conditioned value along execution trajectories.
    Local high-value frames are aligned with semantically meaningful task-progress key states, such as placing the bowl on the plate, opening the drawer, or moving the plate toward the stove.
    This qualitative evidence suggests that the learned value provides dense progress signals beyond final task success.
    }
    \label{fig:value_analysis}
\end{figure}

\paragraph{Value curve analysis.}
To verify whether the value function captures meaningful goal progress, we visualize the learned value estimates along representative trajectories. As shown in Fig.~\ref{fig:value_analysis}, the learned value curves exhibit clear local high-value regions along the execution trajectory. 
Interestingly, these high-value turning points are well aligned with semantically meaningful task-progress key states, such as approaching the target object, establishing a stable grasp, opening the drawer, or placing the object at the goal location. 
This suggests that the learned value function does not merely estimate final task success, but also captures intermediate progress toward the goal. 
By assigning higher values to states that are closer to reachable task-progress milestones, our method provides a progress-aware optimization signal for offline policy improvement, supporting key-state reachability learning from offline demonstrations.

\section{Hyperparameter Analysis}
\label{app:hyperparameter}

In this section, we analyze the sensitivity of our method to the objective weights in Eq.~\ref{eq:total_loss}. 
The final policy objective contains three terms: supervised fine-tuning on achieved goal segments, advantage-guided offline reinforcement fine-tuning, and negative suppression on failed goal segments. 
Their relative importance is controlled by \(\lambda_{\mathrm{SFT}}\), \(\lambda_{\mathrm{RL}}\)respectively:
\[
\mathcal{L}_{\mathrm{total}}(\theta)
=
\lambda_{\mathrm{SFT}}\mathcal{L}_{\mathrm{SFT}}(\theta)
+
\lambda_{\mathrm{RL}}\mathcal{L}_{\mathrm{RL}}(\theta)
.
\]
Unless otherwise specified, we use \(\lambda_{\mathrm{SFT}}=0.4,\lambda_{\mathrm{RL}}=0.6.\)

\begin{table}[t]
\centering
\caption{
Sensitivity analysis of the objective weights. 
The default setting is highlighted in gray. 
}
\label{tab:objective_weight_analysis}
\renewcommand{\arraystretch}{1.08}
\setlength{\tabcolsep}{6pt}
\begin{tabular}{cccccc}
\toprule
\(\lambda_{\mathrm{SFT}}\) 
& \(\lambda_{\mathrm{RL}}\) 
& LIBERO-Goal \\
\midrule
0.1 & 0.9 & \texttt{79.6} \\
0.2 & 0.8 & \texttt{83.2} \\
0.3 & 0.7 &  \texttt{86.8} \\
\rowcolor{gray!15}
0.4 & 0.6 & \texttt{92.3} \\
0.5 & 0.5 & \texttt{89.2} \\
0.6 & 0.4& \texttt{82.5} \\
0.7 & 0.3& \texttt{82.8} \\
\bottomrule
\end{tabular}
\end{table}

\paragraph{Discussion.}
The tow terms play complementary roles. 
The SFT term provides a stable imitation anchor from achieved goal segments. 
The offline RL term further distinguishes actions according to their estimated long-term contribution to goal completion. 
Empirically, we find that using all three components yields more stable and effective policy improvement than relying on any single term alone.

\section{Implementation Details}
\label{app:implementation}

We implement our offline fine-tuning method within the RLinf~\cite{pirl} framework. For fair comparison, our implementation follows the same preprocessing, action normalization, backbone initialization, and optimization setup as the corresponding SFT baseline. The value model is trained before policy fine-tuning and is only used during offline training to compute advantage weights.

\begin{table}[t]
\centering
\caption{Optimization hyperparameters.}
\label{tab:optimization_hyperparams}
\renewcommand{\arraystretch}{1.08}
\setlength{\tabcolsep}{6pt}
\begin{tabular}{lcc}
\toprule
\textbf{Hyperparameter} & \textbf{LIBERO} & \textbf{MetaWorld} \\
\midrule
Optimizer & AdamW & AdamW \\
Policy learning rate & \(2.5\times10^{-5}\) & \(2.5\times10^{-5}\) \\
Value learning rate & \(1.55\times10^{-4}\) & \(1.55\times10^{-4}\) \\
Adam \(\beta_1\) & 0.9 & 0.9 \\
Adam \(\beta_2\) & 0.95 & 0.95 \\
Adam \(\epsilon\) & \(1.0\times10^{-8}\) & \(1.0\times10^{-8}\) \\
Weight decay & \(1.0\times10^{-10}\) & \(1.0\times10^{-10}\) \\
Gradient clipping & 1.0 & 1.0 \\
LR schedule & Cosine warmup & Cosine warmup \\
Warmup steps & 1,000 & 1,000 \\
Policy Training Steps & 30,000 & 10,000 \\
LoRA & Yes & Yes \\
LoRA rank & 32 & 32 \\
\bottomrule
\end{tabular}
\end{table}

\begin{table}[t]
\centering
\caption{Value model training settings.}
\label{tab:value_training_settings}
\renewcommand{\arraystretch}{1.08}
\setlength{\tabcolsep}{6pt}
\begin{tabular}{lcc}
\toprule
\textbf{Setting} & \textbf{LIBERO} & \textbf{MetaWorld} \\
\midrule
Value loss & TD & TD \\
Value training epochs & 5 & 5 \\
Steps per epoch & 600 & 1000 \\
Input images & 2 & 1 \\
Value feature & After VLM & After VLM \\
Pooling mode & Mean token & Mean token \\
Freeze VLM & Yes & Yes \\
Strict checkpoint loading & Yes & Yes \\
Used at inference & No & No \\
\bottomrule
\end{tabular}
\end{table}

\begin{table}[t]
\centering
\caption{Offline objective hyperparameters.}
\label{tab:offline_objective_hyperparams}
\renewcommand{\arraystretch}{1.08}
\setlength{\tabcolsep}{6pt}
\begin{tabular}{lcc}
\toprule
\textbf{Hyperparameter} & \textbf{LIBERO} & \textbf{MetaWorld} \\
\midrule
AWAC temperature \(\lambda\) & 1.0 & 1.0 \\
Maximum advantage weight & 20.0 & 20.0 \\
Normalize advantage & Yes & Yes \\
Advantage clipping & 3.0 & 3.0 \\
External reward model & No & No \\
Online critic model & No & No \\
Online rollouts & No & No \\
\bottomrule
\end{tabular}
\end{table}

\section{Computational Cost Analysis}
\label{app:compute_cost}
We compare the computational cost of our offline fine-tuning method with online RL-based post-training. Unlike online RL methods, our method does not require policy rollout collection during training. All reward annotation, value learning, and policy optimization are performed on the fixed offline dataset. The comparison shows that our offline fine-tuning method is substantially more computationally efficient than online RL-based training. More importantly, our method avoids online environment interaction, making it easier to scale to multiple task suites and safer for robotic manipulation settings where online rollout collection can be expensive or risky.

\begin{table}[H]
\centering
\caption{
Computational cost comparison on LIBERO. 
GRPO denotes the online RL training recipe in RLinf. 
The GRPO cost is reported for one LIBERO task suite, while our method is reported for all four LIBERO suites.
}
\label{tab:compute_cost}
\renewcommand{\arraystretch}{1.12}
\setlength{\tabcolsep}{5pt}
\begin{tabular}{lcccc}
\toprule
\textbf{Method} 
& \textbf{Online Rollouts} 
& \textbf{Hardware} 
& \textbf{Task Suites} 
& \textbf{Wall-clock Time} \\
\midrule
RLinf-GRPO 
& Yes 
& 8$\times$A100 
& 1 LIBERO suite 
& $\sim$175 hours \\
Ours 
& No 
& 4$\times$RTX 4090 48GB 
& 4 LIBERO suites 
& $\sim$68 hours \\
\bottomrule
\end{tabular}
\end{table}

\end{document}